# Qualitative Decision Making Under Possibilistic Uncertainty: Toward more discriminating criteria

**Paul Weng**
LIP6
8 rue du Capitaine Scott
75015 Paris, France

## Abstract

The aim of this paper is to propose a generalization of previous approaches in qualitative decision making. Our work is based on the binary possibilistic utility (PU), which is a possibilistic counterpart of Expected Utility (EU). We first provide a new axiomatization of PU and study its relation with the lexicographic aggregation of pessimistic and optimistic utilities. Then we explain the reasons of the coarseness of qualitative decision criteria. Finally, thanks to a redefinition of possibilistic lotteries and mixtures, we present the refined binary possibilistic utility, which is more discriminating than previously proposed criteria.

## 1 INTRODUCTION

Decision making and more specifically decision making under uncertainty play a central role in AI. The standard approach to modeling such problems is Expected Utility theory (EU). This framework is particularly appealing as it has received much attention and is axiomatically justified (von Neumann and Morgenstern (1944); Machina (1988)). Its application implies that both uncertainty (described by probability) and utility about consequences of actions are numerically valued. Unfortunately in real life problems these conditions are not always fulfilled. Indeed probability may not be suitable to model all kind of uncertainty. Moreover even when it is, probability and utility can be difficult to assess. The first point has led to many generalizations or alternatives to probability theory such as theory of evidence (Shafer (1976)), possibility theory (Dubois and Prade (1990)) or plausibility measures (Friedman and Halpern (1995)). Qualitative knowledge representation and qualitative decision making, which are receiving much interest in the AI community in recent years seem to offer a good framework when facing the difficulties of the second point. In this paper uncertainty is supposed to be represented by possibility theory, which is particularly adapted to modeling situations of partial knowledge.

In qualitative decision making, Boutilier (1994), Bonet and Geffner (1996), Brafman and Tennenholtz (1997) proposed approaches consisting in focusing on best/worst outcomes and/or highest plausibility. More discriminating decision rules have been proposed. Dubois et al. (1998, 2001) axiomatically justified the use of two criteria: optimistic and pessimistic utilities. These two approaches have been recently unified in binary possibilistic utility theory by Giang and Shenoy (2001). Another work of interest is that of Dubois et al. (2000) who proposed an axiomatics for the lexicographic aggregation of pessimistic and optimistic utilities, which consists in applying pessimistic utility first and refining with optimistic utility to break ties. In Giang and Shenoy (2005), a brief discussion is led on the relation beween the latter criterion and binary possibilistic utility. Recently Fargier and Sabbadin (2003) have proposed a refinement of optimistic and pessimistic utilities (replacing min and max by leximin and leximax) and related the refined criteria to EU.

Binary possibilistic utility seems to be a good framework as it is a possibilistic counterpart of EU. However it still suffers from a lack of decisiveness as do many qualitative decision criteria. The aim of this paper is twofold: to give a better understanding of binary possibilistic utility and to provide a more discriminating criterion for decision problems under possibilistic uncertainty. This criterion can be viewed as a generalization of the refinements of optimistic and pessimistic utilities, introduced by Fargier and Sabbadin (2003).

This article is organized as follows: in Section 2 we recall the definition of binary possibilistic utility and its relation with optimistic and pessimistic utilities. Then in Section 3 we provide a new system of axioms for the binary possibilistic utility, which is more similar to usual axiomatizations of EU. In Section 5 we present several sources of the coarseness of binary possibilistic utilities. This motivates the construction of refined binary possibilistic utilities in Section 6 based on a new definition of possibilistic lotteries and mixtures. Finally we conclude in Section 7 by presenting some possible future works.

## 2   BINARY POSSIBILISTIC UTILITY

In qualitative decision making under uncertainty, decisions are represented by functions from a finite set of states to a finite set of consequences. Uncertainty about the actual state is modeled by a possibility distribution over states. Thus each decision is associated with a possibility distribution (or lotteries) over consequences.

This is formalized as follows. Let $X = \{x_1, \ldots, x_n\}$ denote a finite set of consequences. A preference relation $\succsim_X$ is defined over this set. This set has a best element $\overline{x}$ and a worst element $\underline{x}$. Uncertainty is measured on a finite qualitative scale $V$ endowed with an order relation $\geq$. The infinum and the supremum of $V$ are respectively denoted by 0 and 1. The $\max$ and $\min$ operators on $V$ are respectively denoted by $\vee$ and $\wedge$. The order reversing involution in $V$ is denoted by $n$, i.e. $n(0) = 1, n(1) = 0$ and $\lambda \geq \lambda' \Rightarrow n(\lambda) \leq n(\lambda')$.

A simple possibilistic lottery is defined as a function from the set of consequences $X$ to the qualitative scale $V$. The set of all such lotteries is denoted by $\hat{\Pi}(X) = V^X$. Recursively we define the set of compound lotteries that are lotteries over lotteries:

$$\begin{aligned}\hat{\Pi}^1(X) &= \hat{\Pi}(X) \\ \hat{\Pi}^k(X) &= \hat{\Pi}^{k-1}(\hat{\Pi}(X)) \;\; \forall k > 1\end{aligned}$$

The set of all (simple and compound) lotteries is then denoted by $\hat{\Pi}^\infty(X) = \cup_{k=1}^\infty \hat{\Pi}^k(X)$.

A normalized lottery is a lottery such as supremum 1 is reached by at least one consequence. The set of all simple normalized lotteries is denoted by $\Pi(X) = \{\pi \in X^V : \exists x \in X, \pi(x) = 1\}$. Note that $\Pi(X) \subset \hat{\Pi}(X)$. In the same manner, we define the set of all normalized lotteries $\Pi^\infty(X)$. Remark that $\Pi^\infty(X) \subset \hat{\Pi}^\infty(X)$. As the set of consequences is finite, a simple lottery $\pi$ can be written: $[\pi(x_1)/x_1, \ldots, \pi(x_n)/x_n]$. For simplicity, $x$ denotes both an element of $X$ and the degenerated lottery $\pi_x \in \Pi(X)$ such that $\pi_x(x) = 1$ and $\pi_x(z) = 0, \forall z \neq x$.

The following condition of reduction of lotteries is generally implicitly assumed in the works dealing with possibilistic decision making.

$(R)$ (Reduction of lotteries)

$$\forall x \in X, [\lambda_1/\pi_1, \ldots, \lambda_m/\pi_m](x) = \bigvee_{i=1}^m (\lambda_i \wedge \pi_i(x))$$

It allows compound lotteries to be reduced as simple lotteries. Under this condition we have $\Pi(X) = \Pi^\infty(X)$ and $\hat{\Pi}(X) = \hat{\Pi}^\infty(X)$. We will make this condition explicit

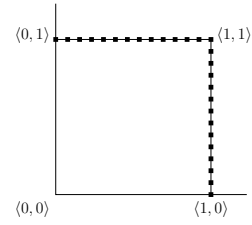

Figure 1: Binary utility scale $U_V$.

in our presentation as in Section 6 this condition will be relaxed.

The decision maker's preference relation over lotteries is denoted by $\succsim$, which reads "at least as good as". Strict preference is denoted by $\succ$ (asymmetric part of $\succsim$) and indifference by $\sim$ (symmetric part of $\succsim$).

In order to formulate the binary possibilistic utility (PU) introduced by Giang and Shenoy (2001) over $\Pi(X)$, we need to define the set of canonical lotteries:

$$\mathcal{C} = \{[\lambda/\overline{x}, \mu/\underline{x}] : \lambda, \mu \in V, \lambda \vee \mu = 1\}$$

This set plays a special role in the definition of PU as its elements define the equivalence classes of $\succsim$. In order for PU to be able to distinguish all these elements, it has to be defined on a rich enough scale. This scale, which is isomorphic to $\mathcal{C}$ is defined as follows:

$$U_V = \{\langle \lambda, \mu \rangle : \lambda, \mu \in V, \lambda \vee \mu = 1\}.$$

Due to constraint $\lambda \vee \mu = 1$, scale $U_V$ is in fact monodimensional (see fig. 1). It is built from the scale of uncertainty $V$. A natural order relation [1] can be defined on $U_V$:

$$\langle \lambda, \mu \rangle \geq_{U_V} \langle \lambda', \mu' \rangle \iff (\lambda \geq \lambda' \text{ and } \mu \leq \mu').$$

Operator $\vee$ is extended as an operator on $U_V \times U_V$ as follows: $\langle \lambda, \mu \rangle \vee \langle \lambda', \mu' \rangle = \langle \lambda \vee \lambda', \mu \vee \mu' \rangle$. Remark that this operator is not the $\max$ operator on $U_V$, which would be define as: $\max(\langle \lambda, \mu \rangle, \langle \lambda', \mu' \rangle) = \langle \lambda \vee \lambda', \mu \wedge \mu' \rangle$.

Operator $\wedge$ is extended as an operator on $V \times U_V$: $\lambda' \wedge \langle \lambda, \mu \rangle = \langle \lambda' \wedge \lambda, \lambda' \wedge \mu \rangle$.

PU is then defined as a function from $\Pi^\infty(X)$ to $U_V$:

$$PU(\pi) = \bigvee_{i=1}^n \bigl(\pi(x_i) \wedge u(x_i)\bigr). \tag{1}$$

where $u : X \to U_V$ assigns a value for each consequence. Function $u$ is named basic utility assignment.

We now present the axioms required in the PU representation theorem:

---
[1] This order is expressed in a simpler form than in Giang and Shenoy (2001). It can be easily checked that the two forms are equivalent.

($B1$) (Preorder) $\succsim$ is reflexive and transitive.
($B2$) (Qualitative monotonicity)
the restriction of $\succsim$ over $\mathcal{C}$ satisfies the following condition:
$[\lambda/\overline{x}, \mu/\underline{x}] \succsim [\lambda'/\overline{x}, \mu'/\underline{x}] \iff \lambda \geq \lambda'$ and $\mu \leq \mu'$.
($B3$) (Substitutability)
$\pi_1 \sim \pi_2 \Rightarrow [\lambda/\pi_1, \mu/\pi] \sim [\lambda/\pi_2, \mu/\pi]$.
($B4$) (Continuity) $\forall x \in X, \exists \sigma \in \mathcal{C}, x \sim \sigma$.

Axiom ($B1$) is slightly weaker than the one used by Giang and Shenoy (2001). Indeed we do not impose the completeness of relation $\succsim$. It can be easily checked that this property results from the other axioms. Axiom ($B2$) expresses a "rational" way to rank the canonical lotteries. When comparing two of these lotteries, one prefers the lottery that gives the greatest possibility to get $\overline{x}$ and the lowest possibility to get $\underline{x}$. Axiom ($B3$) states that equivalent lotteries can be replaced by one another in compound lotteries. Axiom ($B4$) allows the sure consequences to be compared to canonical lotteries. This axiom allows the commensurability of consequences and uncertainty.

The PU representation theorem established in Giang and Shenoy (2001) is stated as follows:

**Theorem 1** *Under Condition* $(R)$, $\succsim$ *on* $\Pi^\infty(X)$ *satisfies Axioms* $(B1)$ *through* $(B4)$ *if and only if there exists a basic utility assignment* $u : X \to U_V$ *such that*

$$\forall \pi, \pi' \in \Pi^\infty(X), \pi \succsim \pi' \iff PU(\pi) \geq_{U_V} PU(\pi').$$

We now present optimistic and pessimistic utilities, axiomatized by Dubois et al. (1998, 2001) and relate them to PU. We assume here that uncertainty and utility are measured on the same scale[2] $V$. In this framework, the basic utility assignment (denoted $v$) is assumed to take values in $V$. We can remark that compared to the PU approach, only half of scale $U_V$ is exploited. Then optimistic and pessimistic utilities are functions from $\Pi^\infty(X)$ to $V$. Optimistic utility writes:

$$U^+(\pi) = \bigvee_{i=1}^n \left(\pi(x_i) \wedge v(x_i)\right) \qquad (2)$$

and pessimistic utility writes:

$$U^-(\pi) = \bigwedge_{i=1}^n \left(n(\pi(x_i)) \vee v(x_i)\right) \qquad (3)$$

where $v : X \to V$.

We introduce two other axioms needed for the representation theorem of optimistic and pessimistic utilities:

($B4^+$) (Optimistic continuity)
$\forall x \in X, \exists \lambda \in V, x \sim [\lambda/\overline{x}, 1/\underline{x}]$.
($B4^-$) (Pessimistic continuity)
$\forall x \in X, \exists \mu \in V, x \sim [1/\overline{x}, \mu/\underline{x}]$.

---
[2]This assumption simplifies the presentation without loss of generality.

The representation theorem for pessimistic utility as formulated by Giang and Shenoy (2001) states as follows:

**Theorem 2** *Under Condition* $(R)$, $\succsim$ *on* $\Pi^\infty(X)$ *satisfies Axioms* $(B1)$ *through* $(B3)$ *and* $(B4^-)$ *if and only if there exists a basic utility assignment* $v : X \to V$ *such that*

$$\forall \pi, \pi' \in \Pi^\infty(X), \pi \succsim \pi' \iff U^-(\pi) \geq U^-(\pi').$$

The relation with PU can be written as follows $U(\pi) = \langle 1, n(U^-(\pi)) \rangle$. The same representation theorem for optimistic utility can be stated with Axiom $(B4^+)$ and we would have $U(\pi) = \langle U^+(\pi), 1 \rangle$.

## 3 A NEW AXIOMATIZATION

As stated by Giang and Shenoy (2001), PU is very similar to expected utility (EU) where $\vee$ is replaced by $+$ and $\wedge$ is replaced by $\times$. This similarity can be axiomatically explained. Indeed depending on how uncertainty is represented, a same set of axioms yields both criteria. Here we propose a set of axioms, more similar to usual axiomatizations of EU:

($C1$) (Total preorder) $\succsim$ is reflexive, transitive, complete.
($C2$) (Non triviality)
$[\lambda/\overline{x}, \mu/\underline{x}] \sim [\lambda'/\overline{x}, \mu'/\underline{x}] \Rightarrow (\lambda = \lambda'$ and $\mu = \mu')$
($C3$) (Weak independence)
$\pi_1 \succsim \pi_2 \Rightarrow [\lambda/\pi_1, \mu/\pi] \succsim [\lambda/\pi_2, \mu/\pi]$.
($C4$) (Continuity)
$\pi_1 \succ \pi_2 \succ \pi_3 \Rightarrow \exists \lambda, \mu$ s.t. $\lambda \vee \mu = 1, [\lambda/\pi_1, \mu/\pi_3] \sim \pi_2$.

Axiom ($C2$) precludes the trivial case where all canonical lotteries are equivalent. The other axioms are inspired by the set of axioms presented by Machina (1988). Axiom ($C3$) states that in a compound lottery replacing a sublottery by a preferred one cannot worsen that lottery. Axiom ($C4$) states that any two lotteries can be combined to be equivalent to any other lottery in between them. These axioms are well-known in decision theory and focus on the way that the decision maker constructs his preferences. From this point of view, these axioms are more natural and give a better understanding of the properties of PU. Indeed the order induced by Axiom ($B2$) is not artificially imposed in this set of axioms. It will result from them.

We now present a lemma that will be useful for the proof of our PU representation theorem.

**Lemma 1** *Under Condition* $(R)$, *if Axiom* $(C3)$ *is verified then for all* $a, b, c, d \in V, a \vee b = c \vee d = 1$, *we have* $(a \geq c$ *and* $b \leq d) \Rightarrow [a/\overline{x}, b/\underline{x}] \succsim [c/\overline{x}, d/\underline{x}]$.

**Proof.** Let $a, b, c, d \in V$ such that $a \vee b = c \vee d = 1$ and $a \geq c$ and $b \leq d$.
By ($C1$), we know that $\overline{x} \succsim \underline{x}$. By ($C3$), $[\lambda/\overline{x}, \mu/\pi] \succsim [\lambda/\underline{x}, \mu/\pi], \forall \pi$. Choose $\pi = [\alpha/\overline{x}, \beta/\underline{x}]$. Then by $(R)$,

$$[\lambda \vee (\mu \wedge \alpha)/\overline{x}, \mu \wedge \beta/\underline{x}] \succsim [\mu \wedge \alpha/\overline{x}, \lambda \vee (\mu \wedge \beta)/\underline{x}] \qquad (4)$$

Case 1: $c = 1$. Then $a = 1$. Let $\mu = \alpha = 1$, $\lambda = d$ and $\beta = b$. Injecting these values in Equation 4 gives the result.
Case 2: $b = 1$. Then $d = 1$. Let $\mu = \beta = 1$, $\lambda = a$ and $\alpha = c$ and the result follows as in the previous case.
Case 3: $a = 1 > c$ and $d = 1 > b$. We get the result from $\lambda = \mu = 1$, $\beta = b$ and $\alpha = c$.
Consequently, $[a/\overline{x}, b/\underline{x}] \succsim [c/\overline{x}, d/\underline{x}]$. ■

We now state and prove our representation theorem.

**Theorem 3** *Under Condition* $(R)$, $\succsim$ *on* $\Pi^\infty(X)$ *satisfies Axioms* $(C1)$ *through* $(C4)$ *if and only if there exists a basic utility assignment* $u : X \to U_V$ *such that*

$$\forall \pi, \pi' \in \Pi^\infty(X), \pi \succsim \pi' \iff PU(\pi) \geq_{U_V} PU(\pi').$$

**Proof.** Under $(R)$, $\Pi^\infty(X) = \Pi(X)$. Therefore we just need to consider simple lotteries.
($\Rightarrow$) Assume $\succsim$ satisfies Axioms $(C1)$ to $(C4)$.
To prove this result, we show that Axioms $(C1)$ through $(C4)$ imply Axioms $(B1)$ through $(B4)$. Obviously $(C1)$ implies $(B1)$, $(C3)$ implies $(B3)$ and $(C4)$ implies $(B4)$.

We now show that the order $U_V$ or equivalently on $\mathcal{C}$ defined by Axiom $(B2)$ can be recovered.

Lemma 1 gives the 'only if' part of Axiom $(B2)$. Now choose $a, b, c, d$ such that $a \vee b = c \vee d = 1$ and $[a/\overline{x}, b/\underline{x}] \succsim [c/\overline{x}, d/\underline{x}]$. Assume $a \neq c$ or $b \neq d$. Otherwise the two lotteries are trivially equivalent. Then by Axiom $C2$, $[a/\overline{x}, b/\underline{x}] \succ [c/\overline{x}, d/\underline{x}]$.
Case 1: $c = 1$. Assume $a < 1$. Then $b = 1 \geq d$. By Lemma 1, we have $[c/\overline{x}, d/\underline{x}] \succsim [a/\overline{x}, b/\underline{x}]$. Thus there is a contradiction and $a = 1$. Similarly if $b \geq d$, there would be a contradiction. Then we have $b < d$.
Case 2: $c < 1$. In the same manner, we prove that $a \geq c$ and $b \leq d$.
This finishes the proof of the 'if' part of Axiom $(B2)$.

($\Leftarrow$) Assume function $PU$ is defined such that $\pi \succsim \pi'$ iff $PU(\pi) \geq_{U_V} PU(\pi'), \forall \pi, \pi' \in \Pi(X)$.
From Theorem 1 relation $\succsim$ verifies Axioms $(B1)$ through $(B4)$, which gives $(C1)$ and $(C2)$.
We first show Axiom $(C4)$. Let $\pi_1, \pi_2, \pi_3$ be three lotteries such that $\pi_1 \succ \pi_2 \succ \pi_3$. For $i = 1 \ldots 3$ it exists $\lambda_i, \mu_i$ such that $\lambda_i \vee \mu_i = 1$ and $PU(\pi_i) = \langle \lambda_i, \mu_i \rangle$. This means that for $i = 1 \ldots 3$, $\pi_i \sim [\lambda_i/\overline{x}, \mu_i/\underline{x}]$. Moreover we know that for $i = 1, 2$, $\lambda_i \geq \lambda_{i+1}$ and $\mu_i \leq \mu_{i+1}$ by Axiom $(B2)$. Let $\pi = [\lambda_2/\pi_1, \mu_2/\pi_3]$. Then by $(B3)$ and $(R)$, $PU(\pi) = \langle \lambda, \mu \rangle = \langle \lambda_2 \vee (\mu_2 \wedge \lambda_3), \mu_2 \vee (\mu_1 \wedge \lambda_2) \rangle$.
Case 1: $\lambda_2 = 1$. Then $\lambda = 1$ and $\mu = \mu_2$.
Case 2: $\mu_2 = 1$. Then $\lambda = \lambda_2$ and $\mu = 1$. Consequently $\pi \sim \pi_2$. This proves $(C4)$.
We now prove Axiom $(C3)$. Let $\pi_1$ and $\pi_2$ be two lotteries such that $\pi_1 \succsim \pi_2$. Then for $i = 1, 2$ it exists $\lambda_i, \mu_i$ such that $\lambda_i \vee \mu_i = 1$ and $PU(\pi_i) = \langle \lambda_i, \mu_i \rangle$. Let $\pi$ be an arbitrary lottery such that $PU(\pi) = \langle \lambda, \mu \rangle$. Let $\alpha, \beta$ such that $\alpha \vee \beta = 1$. Then by $(B3)$ and $(R)$,

$PU([\alpha/\pi_1, \beta/\pi]) = \langle (\alpha \wedge \lambda_1) \vee (\beta \wedge \lambda), (\alpha \wedge \mu_1) \vee (\beta \wedge \mu) \rangle$.
By assumption, we know that $\lambda_1 \geq \lambda_2$ and $\mu_1 \leq \mu_2$. Thus $PU([\alpha/\pi_1, \beta/\pi]) \geq \langle (\alpha \wedge \lambda_2) \vee (\beta \wedge \lambda), (\alpha \wedge \mu_2) \vee (\beta \wedge \mu) \rangle$.
$(C3)$ is then proved. ■

Axioms $(C1)$ through $(C4)$ expressed in a probabilistic setting would entail the use of expected utility.

## 4 REFINED RANKINGS

In this section we make the relation explicit between PU and the lexicographic aggregation of pessimistic and optimistic utilities, which has been axiomatized by Dubois et al. (2000). A discussion on this point has been led by Giang and Shenoy (2005). We give here some new insights.

In this section, we do not limit ourself to normalized lotteries. We define the set of all (non necessarily normalized) canonical lotteries:

$$\hat{\mathcal{C}} = \{[\lambda/\overline{x}, \mu/\underline{x}] : \lambda, \mu \in V\}.$$

Note that constraint $\lambda \vee \mu = 1$ is not imposed anymore. Then an extended utility scale $\hat{U}_V$ can be constructed as follows:

$$\hat{U}_V = \{\langle \lambda, \mu \rangle : \lambda, \mu \in V\}.$$

We endow this scale with the following order relation $>^-$:
$\langle \lambda, \mu \rangle >^- \langle \lambda', \mu' \rangle \iff \mu < \mu'$ or ($\mu = \mu'$ and $\lambda > \lambda'$).
Relation $>^-$ can be seen as the refinement of the second order relation (over $\mu$) by the first order relation (over $\lambda$).

The associated axiom is:
$(D2^-)$ (Pessimistic qualitative monotonicity)
$\succsim$ restricted over $\hat{\mathcal{C}}$ satisfies the following condition:

$$[\lambda/\overline{x}, \mu/\underline{x}] \succ [\lambda'/\overline{x}, \mu'/\underline{x}] \iff \mu < \mu' \text{ or}$$
$$(\mu = \mu' \text{ and } \lambda > \lambda').$$

This axiom is pessimistic in the sense that the possibility of obtaining $\underline{x}$ has more importance than the possibility of getting $\overline{x}$.

Now consider the following axiom:

$(D4)$ (Generalized continuity) $\forall x \in X, \exists \sigma \in \hat{\mathcal{C}}, x \sim \sigma$.

Axiom $(D4)$ states that utilities of consequences can take any value inside the square delimited by $\langle 0, 0 \rangle, \langle 0, 1 \rangle, \langle 1, 1 \rangle$ and $\langle 1, 0 \rangle$ (see fig. 1). The representation theorem exploiting this extended utility scale can be stated as follows:

**Theorem 4** *Under Condition* $(R)$, $\succsim$ *on* $\hat{\Pi}^\infty(X)$ *satisfies Axioms* $(B1)$, $(D2^-)$, $(B3)$ *and* $(D4)$ *if and only if there exists a basic utility assignment* $u : X \to \hat{U}_V$ *such that*

$$\forall \pi, \pi' \in \hat{\Pi}^\infty(X), \pi \succ \pi' \iff PU(\pi) >^- PU(\pi').$$

**Proof.** The proof is essentially similar to that of Theorem 2 of Giang and Shenoy (2001). For lack of space, we only sketch the proof for the 'if' part. First we prove that for any lottery there exists a unique canonical lottery equivalent to it. Under $(R)$, $\Pi^\infty(X) = \Pi(X)$. Therefore we just need to consider simple lotteries. Let $\pi = [\pi_1/x_1, \ldots, \pi_n/x_n]$ be a simple lottery. By $(D4)$, $\forall i, \exists \sigma_i \in \hat{\mathcal{C}}, x_i \sim \sigma_i$. Let $\sigma_i = \langle \lambda_i/\overline{x}, \mu_i/\underline{x} \rangle$ for $1 \leq i \leq n$. By $(B3)$, $\pi \sim \pi' = [\pi_1/\sigma_1, \ldots, \pi_n/\sigma_n]$. Lottery $\pi'$ has only two possible outcomes $\overline{x}$ and $\underline{x}$. Under $(R)$, $\pi' = [\bigvee_{i=1}^n (\pi_i \wedge \lambda_i)/\overline{x}, \bigvee_{i=1}^n (\pi_i \wedge \mu_i)/\underline{x}]$. This defines a unique function $PU(\pi) = \langle \bigvee_{i=1}^n (\pi_i \wedge \lambda_i), \bigvee_{i=1}^n (\pi_i \wedge \mu_i) \rangle$. As $>^-$ and $(D2^-)$ have been defined altogether, the order induced by $(D2^-)$ and $>^-$ are identical. ∎

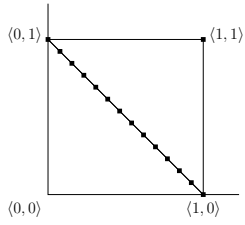

Figure 2: Scale for the lexicographic aggregation of $U^-$ and $U^+$.

We could change Axiom $(D4)$ to Axiom $(D4r)$, restricting consequences to be measured on the segment $\langle 0,1 \rangle$ to $\langle 1,0 \rangle$ (see fig. 2):

$(D4r)$ (Restricted continuity)
$\forall x \in X, \exists \lambda \in V, x \sim [\lambda/\overline{x}, n(\lambda)/\underline{x}]$.

Using Axiom $(D4r)$, the following theorem gives an axiomatization of the lexicographic aggregation of pessimistic and optimistic utilities as a single decision criterion for a single decision maker. Remark that we start with a basic utility assignment valued in $V$ (monodimensional) and we obtain at the end a binary utility (bidimensional).

**Theorem 5** *Under Condition $(R)$, $\succsim$ on $\hat{\Pi}^\infty(X)$ satisfies Axioms $(B1)$, $(D2^-)$, $(B3)$ and $(D4r)$ if and only if there exists a basic utility assignment $v : X \to V$ such that for all $\pi, \pi'$ in $\hat{\Pi}^\infty(X)$:*

$$\pi \succ \pi' \Leftrightarrow \langle U^+(\pi), n(U^-(\pi)) \rangle >^- \langle U^+(\pi'), n(U^-(\pi')) \rangle.$$

**Proof.** From Theorem 4, if Axioms $(B1)$, $(D2^-)$, $(B3)$ and $(D4r)$ are satisfied then it exists $u : X \to \hat{U}_V$ such that $\pi \succ \pi'$ iff $PU(\pi) >^- PU(\pi'), \forall \pi, \pi' \in \Pi(X)$. By $(D4r)$, it exists $v : X \to V$ defined by $\forall x \in X, u(x) = \langle v(x), n(v(x)) \rangle$. Then we can easily check that $PU(\pi) = \langle U^+(\pi), n(U^-(\pi)) \rangle$.

Now if we have $v : X \to V$ that defines two utility functions $U^+$ and $U^-$, we can define $PU(\pi) = \langle U^+(\pi), n(U^-(\pi)) \rangle$. We have obviously Axiom $(D4r)$ by assumption. From Theorem 4, Axioms $(B1)$, $(D2^-)$ and $(B3)$ are verified. ∎

Replacing Axiom $(D2^-)$ by one of the following axioms and defining an order relation associated to it provide results similar to Theorem 4 and 5:

$(D2^+)$ (Optimistic qualitative monotonicity)
$\succsim$ restricted over $\hat{\mathcal{C}}$ satisfies the following condition:

$[\lambda/\overline{x}, \mu/\underline{x}] \succ [\lambda'/\overline{x}, \mu'/\underline{x}] \iff \lambda > \lambda'$ or $(\lambda = \lambda'$ and $\mu < \mu')$.

$(D2^=)$ (Neutral qualitative monotonicity)
$\succsim$ restricted over $\hat{\mathcal{C}}$ satisfies the following condition:

$[\lambda/\overline{x}, \mu/\underline{x}] \succ [\lambda'/\overline{x}, \mu'/\underline{x}] \iff (\lambda > \lambda'$ and $\mu \leq \mu')$ or $(\lambda \geq \lambda'$ and $\mu < \mu')$.

Remark that the order defined by $(D2^=)$ is only partial. Furthermore all these axioms boil down to $(B2)$ when $(D4)$ is restricted to $(B4)$.

## 5 LIMITATIONS OF THESE APPROACHES

The binary possibilistic utility is an improvement over the optimistic and pessimistic utility in terms of discrimination power. However, it can not still differentiate certain simple situations.

**Example 1** *Let $X = \{x_1, \ldots, x_5\}$. These elements are ordered in decreasing order. Then $\overline{x} = x_1$ and $\underline{x} = x_5$. Let $V = \{0, 0.1, 0.2, 0.3, 0.4, 0.5, 0.6, 0.7, 0.8, 0.9, 1\}$. Define the basic utility assignments as: Suppose that we have to*

Table 1: Basic utility assignments

| $u(x_1)$ | $u(x_2)$ | $u(x_3)$ | $u(x_4)$ | $u(x_5)$ |
|---|---|---|---|---|
| $\langle 1, 0 \rangle$ | $\langle 1, 0.1 \rangle$ | $\langle 1, 1 \rangle$ | $\langle 0.1, 1 \rangle$ | $\langle 0, 1 \rangle$ |

*choose between two alternatives: the first one gives for sure gain $x_2$, which is lottery $[1/x_2]$ and the second has two equally possible outcomes $x_1$ and $x_2$, which is represented by lottery $[1/x_1, 1/x_2]$. Then if we compute the PU values of these alternatives ($\langle 1, 0.1 \rangle$), we deduce that they are equivalent, which is quite counterintuitive as the second alternative seems to have a better outlook.*

*Symmetrically we have the same problem with lotteries $[1/x_4, 1/x_5]$ and $[1/x_4]$.*

A comparison between the number of equivalence classes and the number of lotteries gives a first answer to this limitation. The binary possibilistic utility has $2|V| - 1$ equivalence classes. The number of lotteries with $k$ outcomes with possibility 1 and all the other outcomes with possibility strictly less than 1 is $\binom{|X|}{k}(|V| - 1)^{|X|-k}$ denotes the number of . Then we count $\sum_{k=1}^{|X|} \binom{|X|}{k}(|V| - 1)^{|X|-k}$ different lotteries in the whole set $\Pi(X)$. Thus the number of lotteries greatly outnumbers the number of equivalence classes. That partly explains the coarseness of PU.

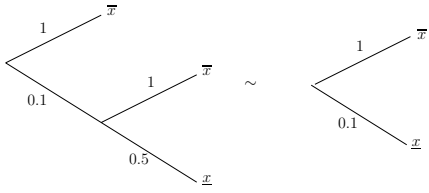

Figure 3: Reduction of compound lotteries.

Moreover an attentive look at the equivalence classes reveals that the situation is even worse. Indeed theses classes are not uniformly distributed. The number of elements in the classes $\langle 1, \lambda \rangle$ and $\langle \lambda, 1 \rangle$ increases as $\lambda$ increases. In the classes $\langle 1, 0 \rangle$ and $\langle 0, 1 \rangle$, there is only one element, respectively $[1/\overline{x}]$ and $[1/\underline{x}]$. The most populated equivalence class is then $\langle 1, 1 \rangle$. This value is reached for a lottery as soon as consequence of utility $\langle 1, 1 \rangle$ has possibility 1 or consequences of utility $\langle 1, \lambda \rangle$ and $\langle \mu, 1 \rangle$ are equally plausible with possibility 1. These consequences hide all other outcomes thus making equivalent all such lotteries.

Approaches like those presented in Section 4 could barely give a better answer as the maximum number of equivalence classes is $|V|^2$. Another proposal due to Fargier and Sabbadin (2003) is to introduce leximin and leximax in place of operators $\wedge$ and $\vee$ to obtain refined criteria. The content of Section 6, also based on leximin and leximax, can be seen as a generalization of this approach in two directions. First, our utility is binary, and second, thanks to a new definition of possibilistic lotteries and mixtures, we can distinguish more finely compound lotteries. Our lotteries will be redefined on richer structures, which will allow infinite equivalence classes.

## 6 REFINED BINARY POSSIBILISTIC UTILITY

Before going into the detail of our formalism, we give an intuition of the reasons why we would gain from a new definition of lotteries. Under Condition $(R)$, those two lotteries are equivalent (see fig. 3): $[1/\overline{x}, 0.1/[1/\overline{x}, 0.5/\underline{x}]]$ and $[1/\overline{x}, 0.1/\underline{x}]$. We think that under $(R)$, compound lotteries are oversimplified. Indeed the simple lottery contains much less information than the compound one. The simple lottery states that only two outcomes are possible: $\overline{x}$ and $\underline{x}$ with respective possibilities 1 and 0.1. Whereas the compound one can be interpreted as follows: we can get immediate outcome $\overline{x}$ with possibility 1 or reach a new uncertain situation represented by a possibilistic lottery $[1/\overline{x}, 0.5/\underline{x}]$. In this new uncertain situation, there is still a possibility to get $\overline{x}$. Resolution of uncertainty can be of special importance like in dynamic decision making where resolution of uncertainty can be sequential. That is why we want to formalize the fact that this compound lottery represents a better alternative than the simple one. In this aim, we need to enrich the set on which lotteries are defined and to replace operators $\wedge$ and $\vee$ by new operators so as to keep all the informations encoded in lotteries.

We introduce a few notations that will be useful. Let $X$ be a set. We denote the set of all finite sequences of elements of $X$ by $X^\infty = \cup_{k=1}^\infty X^k$. If $\vdash$ is an order relation over $X$, we define $\mathcal{R}_X^\vdash : X^\infty \to X^\infty$ the function that ranks sequences of elements of $X$ with respect to $\vdash$, eliminating doubles. If $(x'_1, \ldots, x'_k) = \mathcal{R}_X^\vdash(x_1, \ldots, x_l)$ then $k \leq l$, $\forall i \leq k$, $x'_i \vdash x'_{i+1}$ and $\{x'_1, \ldots, x'_k\} = \{x_1, \ldots, x_l\}$. When there is no risk of confusion, we will drop the subscript and simply write $\mathcal{R}^\vdash$ for $\mathcal{R}_X^\vdash$. An equivalence relation $=$ on $X$ can be naturally extended on $X^\infty$, i.e. $(x_1, \ldots, x_k) = (x'_1, \ldots, x'_l) \iff (k = l$ and $\forall i \leq k, x_i = x'_i)$. If $f : X \to X$ is a function then $f(X)$ denotes the image of $X$ by $f$, i.e $f(X) = \{y \in X : \exists x \in X, f(x) = y\}$.

With the above notations, the set of all finite sequences of elements of $V$ writes $V^\infty$. Let $V^<$ denote the set of all finite increasingly ordered sequences of elements of $V$, which is equal to $\mathcal{R}^<(V^\infty)$. Note that we have $V \subset V^< \subset V^\infty$. If $\lambda$ is an element of $V^<$ then we write $\lambda_i$ for the $i$-th element of the sequence $\lambda$ with the convention that if $i$ is greater than the length of $\lambda$, $\lambda_i = 1$.

A lexicographic order $>_{lex}$ is defined over $V^<$ as follows: $\lambda >_{lex} \lambda' \iff \exists i, \lambda_i > \lambda'_i$ and $\forall j < i, \lambda_j = \lambda'_j$. The weak order is denoted $\geq_{lex}$. This order is simply the leximin on sequences of $V$.

Let $W$ denote the set of all finite decreasingly ordered sequences of elements of $V^<$. We have $W = \mathcal{R}^{>_{lex}}(V^<)$. Remark that $V \subset W$. This is the enriched set on which lotteries will be defined. If $\alpha$ is an element of $W$ then we write $\alpha_i$ for the $i$-th element of the sequence $\alpha$ with the convention that if $i$ is greater than the length of $\alpha$, $\alpha_i = 0$. Note that sequence $\alpha_i$ is an element in $V^<$. Then $\alpha_{1,1}$ denotes the first element of the first sequence of sequence $\alpha$.

Now we define a refined simple possibilistic lottery as a function $X \to W$. We call $\tilde{\Pi}(X)$ the set of simple refined lotteries. Note that the former definition of possibilistic lottery is compatible with this new definition, i.e. $\Pi(X) \subset \tilde{\Pi}(X)$. The set of all (simple or compound) refined lotteries is denoted by $\tilde{\Pi}^\infty(X)$ and we have $\Pi^\infty(X) \subset \tilde{\Pi}^\infty(X)$.

The counterparts on $W$ of operators $\vee$ and $\wedge$ on $V$ are respectively denoted by $\triangledown$ and $\triangle$ and are defined as follows:

$$\alpha \triangledown \alpha' = \mathcal{R}^{>_{lex}}(\alpha, \alpha')$$
$$\alpha \triangle \alpha' = \mathcal{R}^{>_{lex}}((\mathcal{R}^<(\alpha_i, \alpha'_j))_{1 \leq i \leq k, 1 \leq j \leq l}) \text{ if } \alpha \neq 0$$
$$= 0 \text{ otherwise}$$

where $\alpha$ (of length $k$) and $\alpha'$ (of length $l$) are two elements of $W$. Operator $\triangledown$ merges two sequences and ranks them with respect to $>_{lex}$. Then $\big(1, (0.5, 1)\big) \triangledown \big(1, (0.5, 0.6)\big) = \big(1, (0.5, 1), (0.5, 0.6)\big)$. Note that $\mathcal{R}^{>_{lex}}$ deletes doubles. Operator $\triangle$ merges each element of the first sequence with each element of the second sequence and ranks them with respect to $<$. Finally the overall result is ranked with respect to $>_{lex}$. For instance $\big(1,(0.5,1)\big) \triangle \big(1,(0.5,0.6)\big) = \big(1,(0.5,1),(0.5,0.6,1)\big)$.

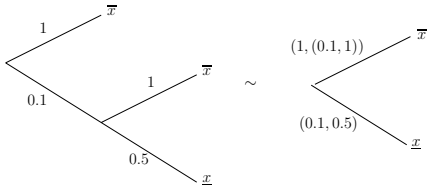

Figure 4: Generalized Reduction of lotteries.

Condition $(R)$ can be rewritten in this setting:
$(RR)$ (Refined reduction of lotteries)

$$[\alpha_1/\pi_1,\ldots,\alpha_m/\pi_m](x) = \bigvee_{i=1}^{m} \alpha_i \triangle \pi_i(x).$$

A look at an example of lottery should clarify these definitions. Suppose we have the following compound lottery (see fig. 4): $[1/\overline{x}, 0.1/[1/\overline{x}, 0.5/\underline{x}]]$. Then this lottery is equivalent to the following refined simple lottery: $[(1,(0.1,1))/\overline{x},(0.1,0.5)/\underline{x}]]$. It is no longer equivalent to $[1/\overline{x}, 0.1/\underline{x}]$. All the information encoded in the compound lottery is retained after reduction. From this example, we remark that reducing lotteries under $(RR)$ does not lose information as under Condition $(R)$.

Under Condition $(RR)$, the set $\Pi^\infty(X)$ can no longer be equated to $\Pi(X)$. In return we have $\tilde{\Pi}^\infty(X) = \tilde{\Pi}(X)$.

We now proceed in the same manner as Giang and Shenoy (2001). We first define the set of refined canonical lotteries as:

$$\mathcal{RC} = \{[\alpha/\overline{x},\beta/\underline{x}] : \alpha,\beta \in W, \alpha_{1,1} \vee \beta_{1,1} = 1\}.$$

Note that this set defines the equivalence classes of our refined criterion and is infinite. From this set we construct the scale on which our utility will be measured. The refined binary possibilistic utility will take values in

$$U_W = \{\langle \alpha,\beta \rangle : \alpha,\beta \in W, \alpha_{1,1} \vee \beta_{1,1} = 1\}.$$

Remark that $U_V \subset U_W$. We need to endow this set with an order relation $>_W^-$:
$\langle \alpha,\beta \rangle >_W^- \langle \alpha',\beta' \rangle \Leftrightarrow$
$\begin{cases} \exists i, (\beta_i <_{lex} \beta'_i \text{ or } (\beta_i = \beta'_i \text{ and } \alpha_i >_{lex} \alpha'_i)) \\ \text{and } \forall j < i, (\alpha_j = \alpha'_j \text{ and } \beta_j = \beta'_j) \end{cases}$
This relation is quite similar to leximax applied on leximin as presented in Fargier and Sabbadin (2003).

Operator $\triangledown$ can be further extended on $U_W \times U_W$:
$\langle \alpha,\beta \rangle \triangledown \langle \alpha',\beta' \rangle = \langle \alpha \triangledown \alpha', \beta \triangledown \beta' \rangle$
Operator $\triangle$ can be further extended on $W \times U_W$:
$\alpha' \triangle \langle \alpha,\beta \rangle = \langle \alpha' \triangle \alpha, \alpha' \triangle \beta \rangle$.

The refined binary possibilistic utility (RPU) is then defined as a function of $\tilde{\Pi}^\infty(X)$ to $U_W$:

$$RPU(\pi) = \bigvee_{i=1}^{n} \pi(x_i) \triangle u(x_i). \qquad (5)$$

We list the axioms that are needed in the representation theorem of RPU:
$(A1)$ (Preorder) $\succsim$ is reflexive and transitive.
$(A2^-)$ (Pessimistic qualitative monotonicity)
$[\alpha/\overline{x},\beta/\underline{x}] \succ [\alpha'/\overline{x},\beta'/\underline{x}]$
$\iff \begin{cases} \exists i, (\beta_i <_{lex} \beta'_i \text{ or } (\beta_i = \beta'_i \text{ and } \alpha_i >_{lex} \alpha'_i)) \\ \text{and } \forall j < i, (\alpha_j = \alpha'_j \text{ and } \beta_j = \beta'_j) \end{cases}$
$(A3)$ (Substitutability)
$\pi_1 \sim \pi_2 \Rightarrow [\alpha/\pi_1,\beta/\pi] \sim [\alpha/\pi_2,\beta/\pi]$.

Axioms $(A1)$ and $(A3)$ are very similar to Axioms $(B1)$ and $(B3)$. In fact the latters are the restriction of the formers to $\Pi^\infty(X)$. Axiom $(A2^-)$ states that lotteries are lexicographically compared following their highest possibilities first. Axiom $(A2^-)$ is pessimistic as the decision maker gives more importance to outcome $\underline{x}$ than to outcome $\overline{x}$.

We obtain the following representation theorem for RPU:

**Theorem 6** *Under Condition $(RR)$, $\succsim$ on $\tilde{\Pi}^\infty(X)$ satisfies Axioms $(A1)$, $(A2^-)$, $(A3)$ and $(B4)$ if and only if there exists a basic utility assignment $u : X \to U_V$ such that $\forall \pi,\pi' \in \tilde{\Pi}^\infty(X), \pi \succsim \pi'$ iff $RPU(\pi) >_W^- RPU(\pi')$.*

**Proof.** The proof is essentially analogous to that of Theorem 2 of Giang and Shenoy (2001). ∎

Similar theorems can be formulated with Axioms $(A2^+)$ or $(A2^=)$ in place of $(A2^-)$ and their associated order relation over $U_W$:
$(A2^+)$ (Optimistic qualitative monotonicity)
$[\alpha/\overline{x},\beta/\underline{x}] \succ [\alpha'/\overline{x},\beta'/\underline{x}]$
$\Leftrightarrow \begin{cases} \exists i, (\alpha_i >_{lex} \alpha'_i \text{ or } (\alpha_i = \alpha'_i \text{ and } \beta_i <_{lex} \beta'_i)) \\ \text{and } \forall j < i, (\alpha_j = \alpha'_j \text{ and } \beta_j = \beta'_j) \end{cases}$
$(A2^=)$ (Neutral qualitative monotonicity)
$[\alpha/\overline{x},\beta/\underline{x}] \succ [\alpha'/\overline{x},\beta'/\underline{x}]$
$\Leftrightarrow \begin{cases} \exists i, ((\alpha_i \geq_{lex} \alpha'_i \text{ and } \beta_i <_{lex} \beta'_i) \\ \text{or } (\alpha_i >_{lex} \alpha'_i \text{ and } \beta_i \leq_{lex} \beta'_i)) \\ \text{and } \forall j < i, (\alpha_j = \alpha'_j \text{ and } \beta_j = \beta'_j) \end{cases}$

Under Axiom $(A2^+)$ the decision maker first focuses on outcome $\overline{x}$ and only after on $\underline{x}$. Finally Axiom $(A2^=)$ gives the same importance to both outcomes. Remark that Axiom $(A2^=)$ only yields a partial order over lotteries. The order relation defined by Axioms $(A2^+)$, $(A2^-)$ or $(A2^=)$ are all compatible with the order relation defined by Axiom $(B2)$.

**Example 2** *Lotteries in $\Pi^\infty(X)$ can now be compared more finely thanks to this new framework as RPU is very discriminating. Indeed consider Example 1 again. We had to compare two lotteries $[1/x_2]$ and $[1/x_1, 1/x_2]$. We can*

*compute their respective RPU:*
$RPU([1/x_2]) = \langle 1, (0.1, 1) \rangle$
$RPU([1/x_1, 1/x_2]) = \langle (1, 1), (0.1, 1) \rangle$
*Then under Axiom $(A2^+)$, $(A2^-)$ or $(A2^=)$, we have $[1/x_1, 1/x_2] \succ [1/x_2]$, which is intuitively satisfying.*

*Similarly, lotteries $[1/x_4, 1/x_5]$ and $[1/x_4]$ are no longer equivalent. And under Axiom $(A2^+)$, $(A2^-)$ or $(A2^=)$, we have $[1/x_4, 1/x_5] \prec [1/x_4]$.*

*Finally we present an example involving lotteries whose PU values are $\langle 1, 1 \rangle$. For instance, consider $[1/x_3]$ and $[1/x_2, 1/x_4]$. PU tells us that these lotteries are equivalent. Their RPU are:*
$RPU([1/x_3]) = \langle 1, 1 \rangle$
$RPU([1/x_2, 1/x_4]) = \langle (1, (0.1, 1)), (1, (0.1, 1)) \rangle$
*Then under Axiom $(A2^+)$, $[1/x_3] \prec [1/x_2, 1/x_4]$. Under Axiom $(A2^-)$, $[1/x_3] \succ [1/x_2, 1/x_4]$. And under Axiom $(A2^=)$, $[1/x_3]$ and $[1/x_2, 1/x_4]$ are incomparable.*

## 7 Conclusion

In this paper, we have proposed a new system of axioms (complete preorder, non triviality, weak independence and continuity) for the binary possibilistic utility introduced by Giang and Shenoy (2001). This new axiomatics is more similar to those proposed for EU. We have axiomatically made the relation explicit between PU and the lexicographic aggregation of pessimistic and optimistic utility. Although PU is an improvement over previously proposed decision criteria, it suffers from a lack of discrimination power. We have highlighted the possible sources of this problem and proposed a refined version of PU thanks to a redefinition of possibilistic lotteries and mixtures. This new criterion can more finely discriminate lotteries as we have shown on a few examples.

The exact relation of RPU with the criterion proposed by Fargier and Sabbadin (2003) needs to be clarified. The nature of this relation seems to be the same as that of PU with optimistic and pessimistic utilities. But as there are several ways to define an order relation over $W$, one can prefer to use pessimistic, optimistic or neutral comparisons. In Fargier and Sabbadin (2003), refinements of optimistic and pessimistic utilities can be computed by means of a numerical expected utility. We suspect that RPU can also be computed that way.

Optimistic and pessimistic utilities have been extended in sequential decision problems by Sabbadin (1999). The new axiomatics that we have proposed here for PU underlines that weak independence is verified. As this axiom is crucial in dynamic decision problems, it appears that PU is also consistent in sequential decision problems as established in Perny et al. (2005). As a refinement of PU, RPU is likely to be consistent in a dynamic setting. We leave this for future works.

**Acknowledgements**

The author wish to thank Patrice Perny, Olivier Spanjaard and the anonymous referees for their useful comments and suggestions to improve this paper.